\documentclass[conference,9pt]{IEEEtran}
\IEEEoverridecommandlockouts
\usepackage{cite}
\usepackage{amscd,amsmath,amssymb,amsfonts,latexsym,mathrsfs,amsthm,mathtools}
\usepackage{bm}
\usepackage{algorithm}
\usepackage{algpseudocode}
\usepackage{graphicx}
\usepackage{textcomp}
\usepackage{xcolor}
\usepackage{tikz-cd}
\usepackage{subcaption}
\usepackage{hyperref}
\usepackage[capitalize]{cleveref}
\usepackage{balance}

\newcommand{\V}[1]{\boldsymbol{\mathbf{#1}}}
\newcommand{\given}{\,\vert\,}

\def\vtheta{{\bm{\theta}}}
\newcommand{\veta}{\bm{\eta}}
\newcommand{\y}{{\bf y}}
\newcommand{\x}{{\bf x}}
\newcommand{\X}{{\bf X}}
\newcommand{\vphi}{\bm{\phi}}

\newcommand{\R}{\mathbb{R}}
\newcommand{\D}{\mathcal{D}}
\newcommand{\N}{\mathcal{N}}

\def\vphi{{\bm{\phi}}}
\def\vPhi{{\bm{\Phi}}}
\DeclareMathOperator{\diag}{diag}

\newtheorem{lemma}{Lemma}

\makeatletter
\def\ps@IEEEtitlepagestyle{%
  \def\@oddfoot{\mycopyrightnotice}%
  \def\@oddhead{\hbox{}\@IEEEheaderstyle\leftmark\hfil\thepage}\relax
  \def\@evenhead{\@IEEEheaderstyle\thepage\hfil\leftmark\hbox{}}\relax
  \def\@evenfoot{}%
}
\def\mycopyrightnotice{%
  \begin{minipage}{\textwidth}
  \centering \scriptsize
  Copyright~\copyright~2025 IEEE. Personal use of this material is permitted. Permission from IEEE must be obtained for all other uses, in any current or future media, including\\reprinting/republishing this material for advertising or promotional purposes, creating new collective works, for resale or redistribution to servers or lists, or reuse of any copyrighted component of this work in other works by sending a request to pubs-permissions@ieee.org.
  \end{minipage}
}
\makeatother

\begin{document}

\title{Decentralized Online Ensembles of Gaussian Processes for Multi-Agent Systems
\thanks{
{$^*$ Equal contribution.
\newline
This work was supported by the National Science Foundation (NSF) under Award Number 2212506.
}}
}

\author{\IEEEauthorblockN{Fernando Llorente$^*$, Daniel Waxman$^*$, and Petar M. Djuri\'c}
\IEEEauthorblockA{Department of Electrical and Computer Engineering \\
\textit{Stony Brook University}\\
Stony Brook, New York, USA}\vspace{-2em}}

\maketitle

\begin{abstract}
Flexible and scalable decentralized learning solutions are fundamentally important in the application of multi-agent systems. While several recent approaches introduce (ensembles of) kernel machines in the distributed setting, Bayesian solutions are much more limited. We introduce a fully decentralized, asymptotically exact solution to computing the random feature approximation of Gaussian processes. We further address the choice of hyperparameters by introducing an ensembling scheme for Bayesian multiple kernel learning based on online Bayesian model averaging. The resulting algorithm is tested against Bayesian and frequentist methods on simulated and real-world datasets.
\end{abstract}


\section{Introduction}
Distributed learning problems concern simultaneous and cooperative inference of $N$ agents, which can communicate across a network with a connectivity graph $\mathcal{G}$. This fundamental problem of multi-agent systems has ubiquity in many applications, {including in swarms of autonomous robots (e.g., drones or rovers), federated learning (e.g., smartphones or IoT devices), and smart grids (e.g., optimization of power distribution and consumption)}. In many systems (e.g., arising from social networks or robot swarms), there is no central ``fusion center'' available, and so learning can only proceed by each agent exchanging data with its neighbors. Algorithms that work in this setting are said to be \emph{decentralized}.

Due to its practical importance, decentralized learning is a classical problem in signal processing, with several popular approaches \cite{djuric2018cooperative}. Diffusion strategies, for example, may be used to solve generic least squares problems \cite{sayed2013diffusion}. Alternatively, frequentist learning can be achieved using distributed optimization methods, such as the alternating direction method of multipliers (ADMM) algorithm \cite{boyd2011distributed,gabay1976dual}.
These tools can be used to solve more complex nonlinear problems via kernel learning \cite{bouboulis2017online}. When no particular kernel is known to be best, multiple kernel learning can achieve even better results \cite{shen2019random}, which is also amendable to the distributed setting \cite{hong2021distributed}.

Our goal is to develop analogous tools in the Bayesian setting, which is comparatively underexplored. In particular, we focus on a Bayesian approach to kernel learning, approximating a global Gaussian process (GP) based on aggregating local statistics of each agent. GPs are among the most common tools in Bayesian machine learning \cite{rasmussen2006gaussian}, representing an expressive Bayesian model with deep theoretical ties to kernel machines. Our approach is based on related work on learning Bayesian linear models \cite{wang2015distributed,dedecius2016sequential}, and parallels developments in frequentist distributed estimation by using random feature (RF) approximation \cite{rahimi2007random,lazaro2010sparse}. Juxtaposed to the frequentist case, however, we will use distributed consensus algorithms rather than distributed optimization algorithms.

Because hyperparameter optimization is crucial for GPs, we also introduce an ensembling approach similar to multiple kernel learning. This approach is based on online Bayesian model averaging \cite{raftery1997bayesian} performed by each agent, mirroring recent trends in conventional GP regression \cite{lu2022incremental,waxman2024doebe}. This allows for truly scalable solutions, without the repeated training of hyperparameters in the online setting.

We summarize our contributions as follows: {(\textbf{1})} we introduce a method that is decentralized, federated, and scalable for GP regression, which asymptotically converges to the ``global'' solution; {(\textbf{2})} we connect the method to other online GP regression methods \cite{lu2022incremental,waxman2024doebe,gijsberts2013real}, showing that a move from Kalman filtering to information filtering makes distributed estimation via consensus %
feasible; and {(\textbf{3})} we extend this method to the ensemble setting using online Bayesian model averaging (BMA); %
finally, we show the competitive performance of our method on several real-world datasets.

We will begin our discussion with a brief problem statement (\cref{sec:problem_statement}) and a review of RF-GPs (\cref{sec:rf_gps}). We then introduce our method for distributed inference (\cref{sec:distributed_rf_gps}), first with a fusion center, and then in a fully decentralized setting. A theoretical framework for ensembling is then introduced (\cref{sec:ensembling}), followed by experiments and discussion (\cref{sec:experiments}) and brief concluding remarks (\cref{sec:conclusion}). We make code to use our method and reproduce our experiments available at \url{https://www.github.com/fllorente/DecentralizedOnlineGPs}.

\section{Problem Statement} \label{sec:problem_statement}

We consider the estimation of a nonlinear function $f(\x)$, $\x\in\R^d$ in a multi-agent system defined by a connected and undirected graph $\mathcal{G} = (\mathcal{N},\mathcal{E})$. The system is composed of $N = |\mathcal{N}|$ agents and each agent only communicates with its neighbors, i.e., denoting with $\mathcal{N}_i$ the set of neighbors of agent $i$, $j \in \mathcal{N}_i$ if and only if $(i,j)\in\mathcal{E}$. The communication steps should be of constant size, with up to $L$ rounds of communication per time step $t$.

In this paper, we learn %
 $f(\x)$ with kernel-based regression. Specifically, we consider the Bayesian estimation of RF-GPs, where nonparametric regression on $f(\x)$ reduces to Bayesian linear regression of a parameter vector $\vtheta$. We aim to compute the posterior distribution of $\vtheta$ in an online and fully decentralized fashion.

\section{Random Feature Gaussian Processes} \label{sec:rf_gps}
While distributed GP estimation has been approached from the full kernel perspective \cite{kontoudis2021decentralized,kontoudis2024scalable}, the resulting methods have high computational costs, do not reach consensus, or are not online. Instead of approximate inference of an exact GP, we thus base our approach on the exact inference of an approximate GP, in particular RF-GPs. We will briefly introduce GPs, and then the RF approximation.

\subsection{Gaussian Processes}
Gaussian processes are stochastic processes often used to model functions in Bayesian machine learning \cite{rasmussen2006gaussian}.
When a GP is used as a prior over an unknown function and given an observed $y$, the %
likelihood $p(y \given f(\V{x}))$ is also Gaussian, the resulting model is conjugate, and the posterior predictive distributions can be computed with elementary linear algebra. A GP may be specified with its kernel $\kappa(\V{x}, \V{x}')$, which captures the correlation between the outputs $f(\V{x})$ and $f(\V{x}')$, and its mean function $\mu(\V{x})$. Without loss of generality, the mean function is typically taken to be identically zero.

The main issue with using GPs for regression is that computation of the GP posterior predictive distribution requires the inversion of an $T\times T$ matrix, where $T$ is the number of data points available. This is difficult in the distributed setting and is generally costly for online inference. Both of these problems will be ameliorated by the random feature approximation introduced in the sequel.

\subsection{Random Feature Gaussian Processes}

A simple approach to approximate GPs is deriving principled linear basis expansions whose predictive distributions asymptotically converge to that of a GP. When the GP is stationary (i.e., the kernel $\kappa(\V{x}, \V{x}')$ depends only on the difference $\V{x} - \V{x}'$), a linear basis expansion can be achieved by randomly sampling from the power spectral density (PSD) $S(\V\omega)$ of the kernel \cite{rahimi2007random,lazaro2010sparse}.
The Wiener-Khinchin theorem shows that the PSD $S(\V\omega)$ is the Fourier transform of $k_{\text{SE-ARD}}(\x - \x')$ \cite[Sec. 4.2.1]{rasmussen2006gaussian}. Using the sampled frequencies, one can construct features known as \emph{random Fourier features}.

As an example, consider the popular squared exponential (SE) kernel with automatic relevance detection (ARD),
\begin{equation}
    k_{\text{SE-ARD}}(\x,\x') = \exp\left(-\sum_{d = 1}^D \frac{(x_{d} - x'_d)^2}{\lambda^2_d}\right), \label{eq:k_se_ard}
\end{equation}
where $\lambda_1, \dots, \lambda_D$ are hyperparameters known as \emph{lengthscales}. Since \cref{eq:k_se_ard} is proportional to a Gaussian distribution, so is its PSD, with $S(\V\omega) = \mathcal{N}(\V\omega \given \mathbf{0}_d, \diag{[\lambda_1^{-2}, \cdots, \lambda_D^{-2}]})$.
Using i.i.d. samples $\V\omega_1, \dots, \V\omega_J \sim S(\V\omega)$, random Fourier features may be computed as
{\small
\begin{align*}
    \vphi(\x) = \frac{1}{\sqrt{J}} [\sin(\x^\top \V\omega_1),\cos(\x^\top \V\omega_1),\cdots,\sin(\x^\top \V\omega_J),\cos(\x^\top \mathbf{v}_J)]^\top.
\end{align*}%
}Thus, $k(\x,\x') \approx \vphi(\x)^\top \vphi(\x')$, and the RF-GP approximation of $f(\x)$ reduces to a linear model with parameters $\vtheta \in \R^{2J}$, i.e.,
\begin{align}\label{eq_RF_GP}
    \widehat{f}(\x) = \vphi(\x)^\top\vtheta,
\end{align}
where $\vtheta \sim \N({\bf 0},\sigma_{\theta}^2{\bf I}_{2J})$.
Thus, given a dataset $\D = (\X, \y)$, $\X\in\R^{t\times d}$, $\y\in\R^t$,
the posterior of $\widehat{f}(\x)$ is determined by the posterior of $\vtheta$, which is also Gaussian. With $\vPhi = [\vphi(\x_1) \dots \vphi(\x_t)] \in \R^{2J\times t}$, the posterior of $\vtheta$ is $p(\vtheta \mid \X, \y) = \mathcal{N}(\vtheta \mid \bm{\mu}_c, \bm{\Sigma}_c)$, with
\begin{equation} \label{eq:rff_gp_posterior_quantities}
\bm{\mu}_c = \frac{1}{\sigma_\text{obs}^2}{\bf \Sigma}_c \vPhi \y, \quad {\bf \Sigma}_c = \left( \frac{1}{\sigma_\text{obs}^2}\vPhi \vPhi^\top + \frac{1}{\sigma_\theta^2} {\bf I} \right)^{-1}.
\end{equation}
It will be convenient later to express the above in equivalent information form. To this end, let ${\bf D}_c = {\bf \Sigma}_c^{-1}$ and $\veta_c = \frac{1}{\sigma_\text{obs}^2}\vPhi\y$. By setting ${\bf D}_0$ to the prior precision matrix, i.e., ${\bf D}_0 = \frac{1}{\sigma_\theta^2} {\bf I}$, \cref{eq:rff_gp_posterior_quantities} is rewritten as
\begin{equation} \label{eq:posterior_w_information_quantities}
    \bm{\mu}_c = {\bf D}_c^{-1} \veta_c, \quad {\bf D}_c = \frac{1}{\sigma_\text{obs}^2}\vPhi \vPhi^\top + {\bf D}_0.
\end{equation}

\section{Distributed Bayesian Inference for RF-GPs} \label{sec:distributed_rf_gps}
We show how to perform distributed Bayesian inference using RF-GPs. For clarity, we will begin with calculations assuming the existence of a fusion center in \cref{sec:distributed_w_fc}. We provide an intuitive explanation of the updates of the fusion center from the perspective of the fusion of probability densities in \cref{sec:fusion_perspective} \cite{koliander2022fusion}, before finally dropping the fusion center assumption in \cref{sec:fully_decentralized_rff_gp}.

\subsection{Distributed Computation with Fusion Center} \label{sec:distributed_w_fc}
Our approach to distributed Bayesian inference is based on the linear model given by \cref{eq_RF_GP}. In particular, we will leverage the conditional independence of $y$ given $\V{x}$ to learn a centralized posterior distribution.
Throughout, we will assume that all agents use the same sample random features, so that their basis expansion function $\phi(\V{x})$ is the same. This can be done without communication by fixing a random seed, or using one of many deterministic random features (for example, orthogonal random features \cite{yu2016orthogonal} or quasi-random Fourier features \cite{yang2014quasi}).

\paragraph*{Decomposition into Local Quantities}
The key to the distributed Bayesian computation of our linear model will be to additively decompose the ``centralized'' posterior into ``local'' quantities of each agent. %
This will allow fully decentralized computation in \cref{sec:fully_decentralized_rff_gp}.

To start, we will split the data into $N$ different groups, which denote the %
data available to the respective agents $n$, i.e.,
\begin{equation}
    \vPhi = [\vPhi_1,\dots,\vPhi_N],\quad \y = [\y_1^\top,\dots,\y_N^\top]^\top.
\end{equation}
For each group, we define the local quantities ${\bf P}_n$ and ${\bf s}_n$,
$$
{\bf P}_n = \frac{1}{\sigma^2} \vPhi_n \vPhi_n^\top + \frac{1}{N\sigma_\theta^2}{\bf I}, \quad {\bf s}_n = \frac{1}{\sigma^2} \vPhi_n \y_n,
$$
where $\vPhi_n$ is the feature matrix for agent  \(n\), and \(\y_n\) is the vector of observed outputs for that agent. Using the block structure of $\V\Phi$, the global quantities ${\bf D}_c$ and $\bm{\eta}_c$ may be computed by summing these local quantities,
\begin{align}
    {\bf D}_c &= \sum_{n=1}^{N} {\bf P}_n = \sum_{n=1}^{N}
    \left(\frac{1}{\sigma_\text{obs}^2} \vPhi_n \vPhi_n^\top+ \frac{1}{N\sigma_\theta^2}{\bf I}
    \right), \label{eq:D_c_sum}
    \\
    \veta_c &= \sum_{n=1}^{N} {\bf s}_n = \sum_{n=1}^{N} \frac{1}{\sigma_\text{obs}^2} \vPhi_n \y_n\ . \label{eq:eta_c_sum}
\end{align}
The posterior mean and covariance are then %
computed as in \cref{eq:posterior_w_information_quantities}.

This additive decomposition is further amendable to sequential/incremental learning where updates correspond to adding new terms with each batch of data.
This update process can be repeated as more data become available, without needing to recompute the full posterior from scratch.
Starting with ${\bf P}_{n,0} = \frac{1}{N\sigma^2_{\theta}}{\bf I}$, ${\bf s}_{n,0} = {\bf 0}$, each agent receives a new batch of data and computes
\begin{align}
    {\bf P}_{n,t} &= \frac{1}{\sigma_\text{obs}^2} \vPhi_{n,t} \vPhi_{n,t}^\top\ ,\label{eq_P_and_s:a} \\
{\bf s}_{n,t} &= \frac{1}{\sigma_\text{obs}^2} \vPhi_{n,t} \y_{n,t}\ .\label{eq_P_and_s:b}
\end{align}
In this online scenario, the centralized posterior is updated using the new local quantities
\begin{align}\label{eq_fusion_center_online}
    {\bf D}_{c,t} &= {\bf D}_{c,t-1} + \sum_{n=1}^{N} {\bf P}_{n,t} = \sum_{\tau=0}^t\sum_{n=1}^{N} {\bf P}_{n,\tau},
    \\
    \veta_{c,t} &=
     \veta_{c,t-1} + \sum_{n=1}^N {\bf s}_{n,t}
     = \sum_{\tau=0}^t\sum_{n=1}^{N} {\bf s}_{n,\tau}\ .
\end{align}

\subsection{A Fusion Perspective} \label{sec:fusion_perspective}
To further elucidate the additive decomposition, we provide a statistical perspective based on the fusion of each agent's ``subposterior.'' To do this, we will first divide the contribution of the prior $p(\V\theta)$ to each agent, then show that fusion with a product rule results in exactly the additive updates above. Throughout, we will use the following well-known fact about Gaussian distributions:
\begin{lemma} \label{lemma:product_of_gaussians}
    Let $p_1(\V{z}) = \mathcal{N}(\mu_1, \Sigma_1)$ and $p_2(\V{z}) = \mathcal{N}(\mu_2, \Sigma_2)$ be two Gaussian densities on the quantity $\V{z}$. Then the normalized density $p_3(\V{z}) \propto p_1(\V{z}) \times p_2(\V{z})$ is also given by a Gaussian, with covariance $\Sigma_3 = (\Sigma^{-1}_1 + \Sigma^{-1}_2)^{-1}$ and mean $\mu_3 = \Sigma_3\Sigma_1^{-1}\mu_1 + \Sigma_3\Sigma_2^{-1}\mu_2$.
\end{lemma}
\paragraph*{Dividing the Contribution of the Prior} Let us consider an isotropic normal prior for $\V\theta$, i.e.,
$ p(\V\theta) = \mathcal{N}(\V\theta \given \V{0}, \sigma_\theta^2 I)$.
For the centralized posterior to use the proper prior, we must divide the prior to each agent, such that the product $\prod_{n} p_n(\V\theta)$ is $p(\V\theta)$. We choose $p_n(\V\theta) \propto p(\V\theta)^{1/N},$ so that agents share the same prior.

\paragraph*{The Centralized Posterior}
After dividing the prior into subpriors, the full posterior can be written as a product of subposteriors,
\begin{equation} \label{eq:product_of_subposteriors}
    p(\V\theta \given \mathcal{D}) \propto \prod_{n=1}^N p_n(\V\theta \given \mathcal{D}_n),
\end{equation}
where $\mathcal{D}_n$ is the data available to agent $n$. The subposteriors are formed by combining the subprior $p_n(\V\theta)$ with the likelihood associated with $\mathcal{D}_n$, i.e., $p(\mathcal{D}_n \given \V\theta)$. Thus, each subposterior is given by $p_n(\V\theta \given \mathcal{D}_n) \propto p_n(\V\theta) p(\mathcal{D}_n \given \V\theta).$
Since $p(\V\theta) \propto \prod_n p_n(\V\theta)$ and (assuming conditional independence of $\mathcal{D}_n$ given $\V\theta$) $p(\mathcal{D} \given \V\theta) = \prod_n p(\mathcal{D}_n \given \V\theta)$, the posterior \cref{eq:product_of_subposteriors} is indeed the correct posterior.

\paragraph*{Fusion Using Subposterior Moments}
Since all distributions of interest are Gaussian, we may summarize the fusion results using the first two moments. In particular, each subposterior has a precision matrix and mean given by
\begin{equation}
    \bm{\Sigma}_n = \left(\frac{1}{\sigma_{\V\theta}^2 N} {\bf I} + \frac{1}{\sigma_{\text{obs}}^2} \V\Phi_n \V\Phi_n^\top\right)^{-1} = {\bf D}^{-1}_n,
\end{equation}
\begin{equation}
    \bm{\mu}_n = \bm{\Sigma}_n\left(\frac{1}{\sigma_\text{obs}^2} \V\Phi_n \V y_n \right) = {\bf D}_n^{-1} \V{\veta}_n.
\end{equation}
As seen in \cref{lemma:product_of_gaussians}, the full posterior then has the information quantities given in \cref{eq:D_c_sum,eq:eta_c_sum}. Therefore, the algebraic manipulation of the previous section is exactly that of product fusion.

\subsection{Decentralized Bayesian Inference} \label{sec:fully_decentralized_rff_gp}

In the case of a fully connected network, each agent can compute the exact centralized posterior by acting as the fusion center.
When the network is not fully connected, each agent must gain knowledge of the posterior through its neighbors. Our approach (inspired by \cite{wang2015distributed} and \cite{hlinka2012likelihood}) is for each agent to approximate the summations $\sum_{n=1}^N {\bf P}_{n,t}$ and $\sum_{n=1}^N{\bf s}_{n,t}$, via consensus algorithms.

Consensus is obtained by iteratively averaging over all neighbors to obtain estimates of the global quantities ${\bf P}_{t} = \sum_n {\bf P}_{n, t}$ and ${\bf s}_t = \sum_n {\bf s}_{n, t}$. For simplicity of presentation, we will assume that the graph $\mathcal{G}$ is undirected and use uniform weights in consensus algorithms, but modifications to using, e.g., Metropolis weights in directed graphs is straightforward \cite{xiao2005scheme}. Letting $N_n$ denote the number of neighbors of agent $n$, consensus iteratively obtains the estimates
\begin{equation}
    \Tilde{{\bf P}}^{(\ell+1)}_{n, t} = \frac{1}{N_n + 1} \left(\Tilde{{\bf P}}^{(\ell)}_{n, t} + \sum_{j \in \mathcal{N}_n} \Tilde{{\bf P}}^{(\ell)}_{j, t} \right), \label{eq:P_consensus}
\end{equation}
\begin{equation}
    \Tilde{{\bf s}}^{(\ell+1)}_{n, t} = \frac{1}{N_n + 1} \left(\Tilde{{\bf s}}^{(\ell)}_{n, t} + \sum_{j \in \mathcal{N}_n} \Tilde{{\bf s}}^{(\ell)}_{j, t} \right), \label{eq:s_consensus}
\end{equation}
where the initial values $\Tilde{{\bf P}}_{n,t}^{(0)}$ and $\Tilde{{\bf s}}_{n, t}^{(0)}$ are the local quantities ${\bf P}_{n, t}$ and ${\bf s}_{n, t}$, respectively. Repeating this process for $\ell = 1, \dots, L-1$, each agent finally obtains an approximate of the global posterior by way of the information quantities
\begin{align}
    {\bf D}_{n, t} &= {\bf D}_{n, t-1} + N\Tilde{{\bf P}}_{n,t}^{(L)}, \label{eq:D_n_update} \\
    \veta_{n, t} &= \veta_{n, t-1} + N\Tilde{{\bf s}}_{n, t}^{(L)}. \label{eq:eta_n_update}
\end{align}
As $L$ increases, $\Tilde{{\bf P}}_{n, t}^{(L)}$ and $\Tilde{{\bf s}}_{n,t}^{(L)}$ will converge to ${\bf P}_t$ and ${\bf s}_t$, and by extension, each agent's posterior approximation converges to the global posterior. The decentralized RF-GP algorithm is summarized in \cref{alg:coopbayesopt_known_hypers}.

\paragraph*{Scalability} By using incremental learning, this approach is both scalable and efficient: at each time step, a fusion center must only aggregate a $2J \times 2J$ matrix and $2J \times 1$ vector of each agent. Moreover, updating the posterior requires only the outer product of a $2J \times 1$ vector. Even when prediction is required, the cost is independent of $t$, with $\mathcal{O}(J^3)$ complexity.

Without a fusion center, the only additional computation is a consensus step; the complexity of this step is once again proportional to $J^2$, and in particular, is $\mathcal{O}(J^2 L)$ complexity. Notably, the time complexity does not depend on the number of agents $N$ directly, though it may influence the choice in $L$ depending on the spectral properties of the graph. Therefore, our approach constitutes a scalable streaming algorithm, so long as $L$ is not too large.

\begin{algorithm}[t]
\caption{Decentralized RF-GPs at Agent $n$} \label{alg:coopbayesopt_known_hypers}
\begin{algorithmic}[1]
\State \textbf{Initialization:} $\Tilde{\mathbf{P}}^{(L)}_{n,0}$ and $\Tilde{\V{s}}^{(L)}_{n,0}$ %
; hyperparameters $\sigma_{\V\theta}^2, \ell, \sigma_{\text{obs}}^2$.
\For{$t = 1 : T$}
    \State \textbf{Step 1:} Acquire new data $(\x_{n, t},\y_{n,t})$ and compute ${\bf P}_{n, t}$, ${\bf s}_{n, t}$.
    \State \textbf{Step 2:} Obtain $\Tilde{{\bf P}}^{(L)}_{n, t}$ and $\Tilde{{\bf s}}^{(L)}_{n, t}$ by peforming $L$ iterations of \cref{eq:P_consensus,eq:s_consensus}.
    \State \textbf{Step 3:} Update the states $\mathbf{D}_{n,t}$ and $\mathbf{\eta}_{n,t}$ based on consensus estimates according to \cref{eq:D_n_update,eq:eta_n_update}.
    \State \textbf{Step 4:} Report the predictive estimate $\widehat{p}_n(\vtheta|\D_t) = \mathcal{N}\left({\bf D}^{-1}_{n,t}\veta_{n,t}, {\bf D}^{-1}_{n,t}\right)$.
\EndFor
\end{algorithmic}
\end{algorithm}

\section{An Ensembling Approach} \label{sec:ensembling}

\begin{figure*}[t]
     \centering
     \begin{subfigure}[b]{0.37\textwidth}
         \centering
         \includegraphics[width=0.9\textwidth]{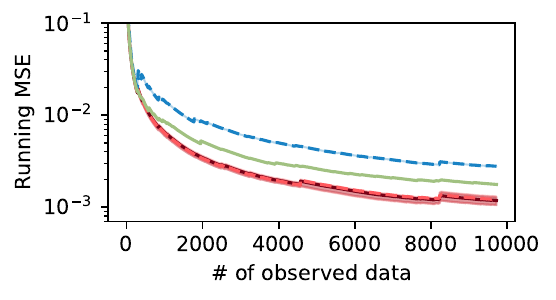}
         \caption{Running Mean Squared Error on Tom's Hardware.}
         \label{fig:single_dataset}
     \end{subfigure}
     \begin{subfigure}[b]{0.5\textwidth}
         \centering
         \includegraphics[width=0.9\textwidth]{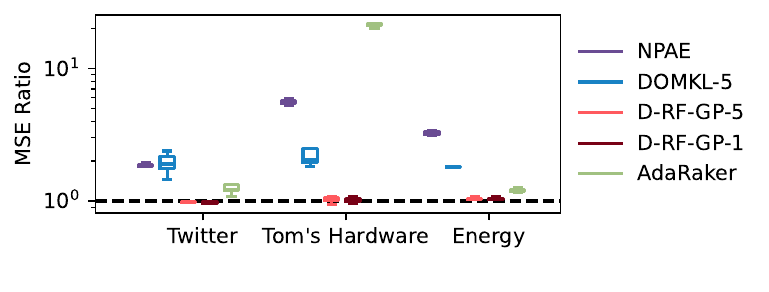}
         \caption{Ratio of MSE w.r.t. SVI GP (lower is better).}
         \label{fig:all_datasets}
     \end{subfigure}
        \caption{The comparative performance of D-RF-GP, DOMKL, AdaRaker, and an SVI GP.}
        \label{fig:results}
\end{figure*}

In this section, we consider decentralized inference of $M$ ensembled and independent models run by each agent.
The use of ensembles is particularly important since we do not consider online learning of the hyperparameters.
In this case, each agent holds $M$ pairs, $(\veta^{(m)}_{n,t}, {\bf D}^{(m)}_{n,t})$ for $m=1,\dots,M$, at each iteration and update them independently following Algorithm \ref{alg:coopbayesopt_known_hypers}.
Then, at each $t$ the agent builds a fused predictive density as a weighted linear mixture of Gaussian distributions
\begin{align}
    \sum_{m=1}^M w^{(m)}_{n,t} \mathcal{N}\left(y_{n,t+1}\mid
    \widehat{y}^{(m)}_{n,t+1}
    ,
    \sigma^{(m),2}_{n,t+1}
    \right),
\end{align}
where $\hat{y}^{(m)}_{n, t+1}$ and $\sigma^{(m), 2}_{n, t+1}$ are determined by the local summary statistics as
\begin{align}
    \widehat{y}^{(m)}_{n,t+1} &= \phi^{(m)}(\x_{n,t+1})^\top
   [{\bf D}^{(m)}_{n,t}]^{-1}\veta^{(m)}_{n,t}, \label{eq_y_hat_agent_model_m} \\
    \sigma^{(m),2}_{n,t+1} &= \phi^{(m)}(\x_{n,t+1})^\top[{\bf D}^{(m)}_{n,t}]^{-1} \phi^{(m)}(\x_{n,t+1}) + \sigma^2_\text{obs}.
\end{align}
The $w^{(m)}_{n,t}$ represents the weight given by agent $n$ to the model $m$ at time $t$, the details of which are the main topic of this section.

\subsection{Bayesian Model Averaging}

One popular way of combining different predictive densities in a Bayesian context is through BMA \cite{raftery1997bayesian}. In BMA, each model is assigned a weight proportional to its marginal likelihood, $p(\y|\x, \mathcal{M}_m)$, where $\mathcal{M}_m$ denotes the $m$-th model.
Using the chain rule, we can decompose
$$
\log p(\y_{1:t}|\mathcal{M}_m) = \sum_{t=1}^T \log p(\y_t| \y_{1:t-1}, \mathcal{M}_m),
$$
where $\y_{1:0}$ is understood to be empty. The takeaway from this decomposition is that we may compute the marginal likelihood recursively by summing the evaluation of the predictive density on unseen data, which we call \emph{online BMA}.

\subsection{The Issue in the Decentralized Setting and a Solution}
In the multi-agent context, the arriving batch $\y_{t+1} = \{y_{1,t+1},\dots,y_{N,t+1}\}$ is spread among the $N$ agents. However, we cannot decompose the evaluation as
\begin{align}
    \log p(\y_{t+1}|\y_{1:t},\mathcal{M}_m) \neq \sum_{n=1}^N\log p(y_{n,t+1}|\y_{1:t},\mathcal{M}_m), \label{eq:bad_marginal_likelihood}
\end{align}
as equality only holds if the batch $\y_{t+1}$ is independent. We therefore present two potential schemes to update the weights.

\paragraph*{Local BMA} The first potential solution is to exclusively perform BMA locally, with each agent using only its private data. This approach is likely to be overly conservative with weights since only $1/N$ of the total data will be used to compute $w_{n,t}^{(m)}$. The corresponding update for $w^{(m)}_{n,t}$ after observing $y_{n,t+1}$ is
\begin{align}\label{eq_bma_agent}
    \log \widetilde{w}^{(m)}_{n,t+1} = \log \widetilde{w}^{(m)}_{n,t} + \log\mathcal{N}\left(y_{n,t+1}\mid
    \widehat{y}^{(m)}_{n,t+1}
    ,
    \sigma^{(m),2}_{n,t+1}
    \right)
\end{align}
where $w^{(m)}_{n,t} = \widetilde{w}^{(m)}_{n,t}/\sum_{j=1}^M\widetilde{w}^{(j)}_{n,t}$.

\paragraph*{Independent Consensus BMA} While equality does not hold in \cref{eq:bad_marginal_likelihood}), this decomposition is often used as an approximation in the (centralized) distributed GP literature \cite{liu2018generalized}.
After assuming each agent has a good approximation to the centralized posteriors $\mathcal{N}([{\bf D}^{(m)}_{c,t}]^{-1}\veta^{(m)}_{c,t},[{\bf D}^{(m)}_{n,t}]^{-1})$ for $m=1,\dots, M$,  a consensus step can be used to compute $\sum_{n=1}^N\log p(y_{n,t+1}|\y_{1:t},\mathcal{M}_m)$. Thus, under the marginal independence assumption, each agent will approximate the same BMA weight as a fusion center by replacing the log-likelihood of \cref{eq_bma_agent} with the consensus estimate of the sum
\begin{align}
\sum_{n=1}^N\log\mathcal{N}\left(y_{n,t+1}\mid
    \widehat{y}^{(m)}_{n,t+1}
    ,
    \sigma^{(m),2}_{n,t+1}
    \right).
\end{align}

\section{Experiments and Discussion} \label{sec:experiments}
To empirically validate our method, we compare our proposed method to both online and batch learners. For online learning, we consider AdaRaker \cite{shen2019random}, a state-of-the-art method for multiple kernel learning (which is centralized and non-Bayesian) and DOMKL \cite{hong2021distributed}, a fully decentralized multiple kernel learning algorithm (which is non-Bayesian). For batch learners, we consider NPAE \cite{rulliere2018nested,kontoudis2021decentralized}, a GP algorithm with distributed training and inference (which is centralized and not online), and the SVI GPs using $100$ inducing points.
Note that NPAE and SVI GPs are both offline algorithms that learn hyperparameters instead of ensembling. With NPAE, in particular, na\"ive online learning is prohibitively expensive. We choose NPAE as it bounds the performance of similar decentralized GP algorithms (DEC-NPAE and DIST-NPAE in \cite{kontoudis2021decentralized}) but with higher communication costs.
We use the authors' codes for NPAE and AdaRaker, and the implementation of SVI GPs from gPyTorch \cite{hensman2013gaussian} and optimize using Adam \cite{Kingma15}.
{For DOMKL, we performed a hyperparameter search and used hyperparameters that provide reasonable results across all datasets.}
We provide results for the Tom's Hardware \cite{kawala2013predictions}, Energy \cite{candanedo2017data}, and Twitter \cite{kawala2013predictions} datasets used in \cite{shen2019random}.

For each experiment, we ran the decentralized algorithms (D-RF-GP-5 and DOMKL-5) with $N=5$ agents, each with three models using SE kernels with lengthscales $\ell \in \{10^{k}\}_{k=-1}^1$. The same SE kernels were used in AdaRaker. For all models, $\sigma^2_{\V\theta} = 1$, $\sigma^2_{\text{obs}} = 10^{-2}$, and $J=50$.
{We also consider 5 agents in NPAE with a single SE kernel.}
The ``independent consensus'' flavor of BMA was chosen, and all consensus algorithms were executed for $L=10$ iterations. We also ran the single-agent version of D-RF-GP (D-RF-GP-1), which is mathematically equivalent to the IE-GP \cite{lu2022incremental} and OE-RFF \cite{waxman2024doebe}.
Random communication graphs with edge probability $0.25$ were chosen, rejecting graphs that were not strongly connected.

To illustrate the online nature of our method, we show the ``running MSE'' of each method on Tom's Hardware in \cref{fig:single_dataset}. For a fair comparison, we compute the MSE every $N_{\text{max}}$ observations so that each MSE is computed after the same number of observations. Thus,
\begin{align*}
    \mbox{MSE}(t) = \frac{1}{N \lfloor  t / N_{\text{max}}\rfloor} \sum_{\tau=1}^{\lfloor t /  N_{\text{max}}\rfloor} \sum_{n=1}^N\left(
    \widehat{y}_{n,N_{\text{max}} \tau} - y_{n,N_{\text{max}}\tau}
    \right)^2,
\end{align*}
where  $y_{n,\tau}$  represents the observation received by agent $n$ and $\widehat{y}_{n,\tau}$ is the prediction using all observations up to $\tau-1$, where $\widehat{y}_{n,\tau} = \sum_{m=1}^Mw^{(m)}_{n,\tau-1}\widehat{y}^{(m)}_{n,\tau}$ for our case, with $\widehat{y}^{(m)}_{n,\tau}$ given by \eqref{eq_y_hat_agent_model_m}.
In NPAE, following \cite{kontoudis2021decentralized} we take $\widehat{y}_{n,\tau} = \widehat{y}_{1,\tau}$ for all $n$ (namely, we use the prediction of agent 1 as the ``aggregated'' prediction of the agents.)

In \cref{fig:all_datasets}, we computed the MSE of each method on a hold-out dataset corresponding to the last 1000 observations of each dataset. To normalize the results, we present the ratio of each method's MSE to that of the SVI GP.
We find surprisingly competitive performance of D-RF-GPs with respect to a GP with centralized training and inference. We also find that D-RF-GP-5 tends to perform similarly to D-RF-GP-1.

\section{Conclusion} \label{sec:conclusion}
We introduced a fully decentralized, consensus-based method for learning Gaussian processes using the random features approximation. We then introduced an ensembling approach for incorporating multiple sets of hyperparameters, and showed its effectiveness on real-world datasets. Future work could incorporate more expressive basis expansions, as shown beneficial by \cite{waxman2024doebe}, incorporate alternative weighting algorithms, or apply this distributed learning to the optimization of an unknown function.

\newpage
\balance

\bibliographystyle{IEEEtran}
\bibliography{refs}

\end{document}